\documentclass[runningheads]{llncs}

\usepackage{times}
\usepackage{epsfig}
\usepackage{graphicx}
\usepackage{amsmath}
\usepackage{amssymb}
\usepackage{booktabs}
\usepackage{multirow}
\usepackage{tabularx}
\usepackage{subfigure}
\usepackage{color, colortbl}
\usepackage{algorithm}
\usepackage{algorithmic}
\usepackage{ifdraft}

\usepackage[width=122mm,left=12mm,paperwidth=146mm,height=193mm,top=12mm,paperheight=217mm]{geometry}
\newcommand{\sota}{\textit{state-of-the-art}}
\newcommand{\etal}{\textit{et al.}}
\newcommand{\cmt}[1]{\ifdraft{ \textcolor{red}{- #1 -} \\}{}}
\newcommand{\punch}{\ifdraft{ \textcolor{red}{\underline{Punchline:}} }{}}

\begin{document}



\pagestyle{headings}

\mainmatter

\def\ECCV14SubNumber{--}  
\title{Feature sampling and partitioning for visual vocabulary generation on large action classification datasets. } 

\author{Michael Sapienza$^1$ \and
        Fabio Cuzzolin$^1$ \and
        Philip H.S. Torr$^2$ }


\institute{   Department of Computing and Communications Technology, 
              Oxford Brookes University. 
           \and 
                Department of Engineering Science,  
                University of Oxford. 
          }





\maketitle

\begin{abstract}

\cmt{Motivation for yet another BoF evaluation paper.}
The recent trend in action recognition is towards larger datasets, 
an increasing number of action classes and larger visual vocabularies. 
State-of-the-art human action classification in challenging video data is currently based on 
a bag-of-visual-words pipeline in which space-time features are aggregated globally to form a histogram. 
The strategies chosen to sample features and construct a visual vocabulary are critical to performance, 
in fact often dominating performance. 
In this work we provide a critical evaluation of various approaches to building a vocabulary and show that good practises do have a significant impact. 
By subsampling and partitioning features strategically, 
we are able to achieve state-of-the-art results on 5 major action recognition datasets using relatively small visual vocabularies. 

\end{abstract}

\section{Introduction}
\label{sec:introduction}
\cmt{Why is human action recognition worth researching?} 
Human action recognition on huge datasets collected from the Internet is an increasingly important research area,
given the popularity of smart-phone videos and online sharing websites. 
The recognition of human actions from challenging data is essential to facilitate the automatic organisation, 
description and retrieval of video data from massive collections \cite{perronnin-cvpr-2010}. 
\cmt{Why is it hard?}
Querying unconstrained video data is hard, 
for online clips are often of low quality, 
and affected by erratic camera motions, 
zooming and shaking. 
Variations in viewpoint and illumination pose additional difficulties. 
Furthermore, human action clips possess a high degree of within-class variability and are inherently ambiguous since the start and end of an action are hard to define. 


\cmt{Current state-of-the-art}
So far the best global representation for challenging video data has been in the form of a bag-of-features (BoF) histogram, 
and its extensions to Fisher vectors \cite{oneata-2013}.  
The surprising success of BoF may be attributed to its ability to summarise local features as a simple histogram, 
without regard for human detection, pose estimation or the location of body-parts, 
which so far cannot be extracted reliably in unconstrained action videos \cite{sapienza-2013}. 
Moreover, the histograms provide a fixed length representation for a variable number of features per video, 
which is ideal for traditional learning algorithms \cite{farquhar-2005}. 

\cmt{Problems with state-of-the-art}
Despite extensive work showing various aspects in which BoF may be improved to squeeze out additional performance 
\cite{csurka-2004,perronnin-2006,nowak-2006,farquhar-2005,jiang-2007,nister-2006,sivic-2003}, 
there remains interesting questions yet to be evaluated empirically, 
particularly when considering huge \emph{action} classification datasets with a large number of classes \cite{kuehne-2011,soomro-2012}. 
 
For example, the recent UCF101 dataset \cite{soomro-2012} contains $101$ action classes and $\approx 13,000$ video clips. 
Using the state-of-the-art Dense Trajectory features, this generates $\sim679$Gb of features. 
Therefore one cannot simply load all the training features into memory. 
Furthermore, randomly subsampling the features generates a bias towards action classes associated with a greater share of videos, 
or disproportionately longer video sequences.

Secondly, state-of-the-art space-time descriptors are typically formed by a number of components. 
The issue arises of whether it is best to learn a single visual vocabulary over the whole, joint feature space, 
or to learn a separate vocabulary for each feature component. 
Finally, visual vocabularies may be learnt separately for each action category \cite{farquhar-2005}. 
Although it generates redundant visual words which are shared among multiple action classes, 
such a strategy is worth exploring for the dimensionality of the representation increases linearly with the number of classes $C$. 
Thus learning multiple dictionaries per-category, 
each with a small number of clusters $K$, may still produce large vector representations since the final dictionary size will be $K \times C$. 

\cmt{Our Proposal to improve things}
In this work we explore various ways of generating visual vocabularies for action classification that address the following unanswered questions: 
i)~What is the best way to randomly subsample features to build a vocabulary? 
ii)~What are the effects of learning separate rather than joint visual vocabularies when considering multiple feature components or multiple action classes? 

\section{Related work}
Current \sota\ human action classification systems rely on the global aggregation of local space-time features \cite{oneata-2013}. 
For example, in the bag-of-features (BoF) approach, 
space-time descriptors are usually clustered into $K$ centroids to form a \textit{visual vocabulary}.
Each video descriptor is then assigned to the nearest cluster in the visual vocabulary,  
and a video clip is represented by a fixed-length histogram,
where each bin represents the frequency of occurring \textit{visual words}. 

The areas in which BoF may be improved were broadly outlined by Nowak \etal\ \cite{nowak-2006} and Jiang \etal\ \cite{jiang-2007}. 
These include the \emph{patch sampling strategy}, 
for which uniform random sampling of the image space was shown to outperform interest point-based samplers such as Harris-Laplace and Laplacian of Gaussian \cite{nowak-2006}. 
Previous action recognition BoF evaluation \cite{wang-2009} focused on comparing a variety of local spatio-temporal features, 
and also found that dense sampling consistently outperformed interest point detection in realistic video settings. 
Later, the authors of \cite{wang-2009} proposed a new \emph{visual descriptor} called Dense Trajectories which achieved the state-of-the-art in action classification \cite{wang-2011}. 
Further areas of improvement were the \emph{feature-space partitioning algorithm}, 
for which Nister \etal\ \cite{nister-2006} extended $k$-means to produce vocabulary trees, and Farquhar \etal\ \cite{farquhar-2005} used a Gaussian mixture model instead of $k$-means; 
the \emph{visual vocabulary size} also improved performance, though it saturated at some data-dependent size \cite{nowak-2006}. 
Another area of improvement was deemed to be the \emph{weighting of frequency components} in a histogram (tf-idf) \cite{sivic-2003}, 
and the use of sets of bipartite histograms representing universal and class-specific vocabularies \cite{perronnin-2006}.
More recently, Fisher and VLAD vectors achieved excellent results by using small vocabularies and soft quantisation instead of a hard assignment to form high dimensional vectors \cite{perronnin-eccv-2010,jegou-2011}. 
A comprehensive evaluation of various feature encoding techniques on image classification datasets was provided by Chatfield \etal\ \cite{chatfield-2011}. 
Finally the choice of \emph{classification algorithm} has largely been dominated by support vector machines \cite{csurka-2004,laptev-2008,wang-2011}. 
In contrast to previous work \cite{nowak-2006,wang-2009,chatfield-2011}, 
here we focus on feature subsampling and partitioning after feature extraction has taken place but prior to encoding.




For earlier systems which used BoF for object retrieval in videos \cite{sivic-2003}, 
building a visual vocabulary from all the training descriptors was known to be a ``gargantuan task'' \cite{sivic-2003}. 
Thus researchers tended to select subsets of the features, dedicating only a sentence to this process. 
For example, quoting \cite{sivic-2003}, ``a subset of 48 shots is selected covering about 10k frames which represent about 10\% of all frames in the movie'', 
amounting to 200k 128-D vectors for learning the visual vocabulary. 
Csurka \etal\ trained a vocabulary with ``$600k$ descriptors'' \cite{csurka-2004}, 
whilst Nister \etal\ used a ``large set of representative descriptor vectors'' \cite{nister-2006}. 
In order to construct visual vocabularies, Jiang \etal\ used all the training keypoint features in the PASCAL dataset, 
and subsampled $80k$ features in the TRECVID dataset \cite{jiang-2007}. 
In the evaluation proposed here, 
we clarify this process by comparing two specific methods to sample features from the smallest to largest action recognition datasets available. 
The sheer volume of features extracted from each dataset used here is listed in Table \ref{tab:feature_stats}.



It is clear that with thousands of videos in the training set, 
in practice one cannot use all of the features to build a vocabulary.  
Also note that in video classification datasets, 
longer sequences have disproportionately more features than shorter ones, 
in contrast to image datasets.   
Moreover, some action classes also have many more videos associated with them than others. 
To see the extent by which the number of features per video varies in action recognition, 
check the difference between the `Max' and `Min' row of Table~\ref{tab:feature_stats}. 
Thus, in general, sampling uniformly at random may result in a pool of features which is biased towards a particular class. 
For example in the Hollywood2 dataset \cite{marszalek-2009} `driving' action clips are typically much longer than `standing up' clips. 
\punch\ We hypothesise that the subsampling strategy may have a significant impact on the classification performance especially when dealing with a large number of action classes and videos. 

\begin{table}[tb]
\caption{Various statistics on the amount of Dense Trajectory features extracted per video from the KTH, Hollywood2 (HOHA2), HMDB and UCF101 datasets. 
The number of action categories are denoted within parenthesis.
The `Sum' and `Mem' are calculated over the whole dataset assuming 32-bit floating point. Following these are the mean, 
standard deviation, median, maximum, and minimum number of features per video in each dataset. }
\begin{center}

\begin{tabular}{|l||c|c|c|c|}
\hline
						& \textbf{KTH (6)} 		& \textbf{HOHA2 (12)} 	& \textbf{HMDB (51)} 	& \textbf{UCF-101} \\ \hline 

Sum {\scriptsize $(\times10^\textbf{6}$)}	& 4 			& 28 			& 99 	 		& 421	\\ \hline

Memory  (GB) 					& 6 			& 45 			& 160 	 		& 679	\\ \hline \hline

Mean {\scriptsize $(\times10^\textbf{3}$)}	& 9			& 34	 		& 16	 		& 32 \\ \hline 

Std Dev {\scriptsize $(\times10^\textbf{3}$)}	& 5 			& 44			& 14 	 		& 30	\\ \hline

Median {\scriptsize $(\times10^\textbf{3}$)}	& 8 			& 20 			& 13 	 		& 22	\\ \hline

Maximum {\scriptsize $(\times10^\textbf{3}$)}	& 36 			& 405 			& 138 	 		& 410	\\ \hline

Minimum						& 967 			& 438 			& 287 	 		& 358	\\ \hline

\end{tabular}

\end{center}
\label{tab:feature_stats}
\end{table}

\cmt{bof per category}
The selection of good partitioning clusters to form a visual vocabulary is also important, 
as they form the basic units of the histogram representation on which a classifier will base its decision \cite{farquhar-2005}. 
Thus, for a categorisation task, having clusters which represent feature patches that distinguish the classes is most likely to make classification easier. 
This motivates per-category clustering, to preserve discriminative information that may be lost by a universal vocabulary \cite{farquhar-2005}, 
especially when distinct categories are very similar. 
The downside is that learning a separate visual vocabulary per class may also generate many redundant clusters when features are shared amongst multiple categories. 
On the other hand, since the complexity of building visual vocabularies depends on the number of cluster centres $K$, 
clustering features independently allows to reduce $K$ whilst keeping the representation dimensionality high, 
and makes the vocabulary learning easily parallelisable.
\punch\ So far, per-category training has seen promise on a single dataset with a small number of classes \cite{farquhar-2005}; 
it is therefore to be seen how it performs on challenging action classification data with a large number of action classes.

\textbf{Contributions.}
First and foremost we demonstrate various design choices that lead to an improved classification performance across 5 action recognition datasets and 3 performance measures.
Secondly, we propose a simple and effective feature sampling strategy used to scale the vocabulary generation to a large number of classes and thousands of videos (c.f. Algorithm \ref{alg:sampling}).
%
The outcomes of this study suggest that 
i) sampling a balanced set of features per class gives minor improvements compared to uniform sampling, 
ii) generating separate visual vocabularies per feature component gives major improvements in performance, 
and iii) BoF per-category gives very competitive performance for a small number of visual words, 
achieving state-of-the-art on KTH with only 32 clusters per category, 
although Fisher vectors won out on the rest.

%
With a simple and clear-cut algorithm, 
the results we obtained on the 5 major datasets in action classification are comparable or outperform the state-of-the-art. 
The code and parameter settings are made available online\footnote{URL for code}.

\section{Datasets and performance measures}
\label{sec:datasets}
The \emph{KTH} dataset \cite{schuldt-2004} contains 6 action classes, 25 actors, and four scenarios. 
The actors perform repetitive actions at different speeds and orientations.
Sequences are longer when compared to those in the YouTube\cite{liu-2009} or the HMDB51\cite{kuehne-2011} datasets,
and contain clips in which the actors move in and out of the scene during the same sequence.
We split the video samples into training and test sets as in \cite{schuldt-2004},
and considered each video clip in the dataset to be a single action sequence.

The \emph{YouTube} dataset \cite{liu-2009} contains videos collected online from 11 action classes. 
The data is challenging due to camera motion, cluttered backgrounds, and a high variation in action appearance, scale and viewpoint. 
In our tests, 1600 video sequences were split into 25 groups, 
following the author's evaluation procedure of 25-fold, leave-one-out cross validation.

The \emph{Hollywood2} dataset \cite{marszalek-2009} contains instances of 12 action classes collected from 69 different Hollywood movies.
There are a total of 1707 action samples each 5-25 seconds long, 
depicting realistic, unconstrained human motion, although the cameras often have their view centred on the actor. 
The dataset was divided into 823 training and 884 testing sequences, as in \cite{marszalek-2009}. 
Note that all videos in this dataset were initially downsampled to half their size. 

The \emph{HMDB} dataset \cite{kuehne-2011} contains 51 action classes, with a total of 6849 video clips collected from movies,
the Prelinger archive, YouTube and Google videos. Each action category has a minimum of 101 clips.
We used the non-stabilised videos with the same three train-test splits proposed by the authors \cite{kuehne-2011}.

The \emph{UCF101} dataset \cite{soomro-2012} contains 101 action classes, $\sim$13k clips and 27 hours of video data. 
This is currently the largest and most challenging video dataset due to the sheer volume of data, 
high number of action classes and unconstrained videos.
We used the recommended three train-test splits proposed by the authors \cite{soomro-2012}. 

To quantify the respective algorithm performance, 
we followed \cite{sapienza-2012} and report the accuracy (Acc), 
mean average precision (mAP), and mean F1 scores (mF1).


\section{Experiments}
In this section we detail the experiments used to answer the questions set out in section \ref{sec:introduction}. 
We keep some parameter settings constant throughout the tests (\S~\ref{subsec:constant_params}) and let others vary (\S~\ref{subsec:variable_params}). 
The specifics of our vocabulary sampling and generation strategy are laid out in Algorithm \ref{alg:sampling}. 

\subsection{Constant experimental settings}
\label{subsec:constant_params}
\textbf{Space-time features. }
Amongst the wide variety of space-time interest point detectors \cite{laptev-2003,willems-2008}
and descriptors \cite{laptev-2008,klaser-2008,jhuang-2007,le-2011},
the Dense Trajectory features of Wang \etal\ \cite{wang-2011} gave the \sota\ results in experiments with challenging action video data \cite{wang-2011,sapienza-2012}.
A Dense Trajectory feature is formed by the concatenation of five feature components: optic flow displacement vectors,
the histogram of oriented gradient and histogram of optic flow (HOG/HOF) \cite{laptev-2008},
and the motion boundary histograms (MBHx/MBHy) \cite{dalal-2006}.
This feature is computed over a local neighbourhood along the optic flow trajectory,
and is attractive because it brings together the local trajectory's shape,
appearance, and motion information \cite{wang-2011}.

These Dense Trajectory features were pre-computed from video blocks of size $32 \times 32$ pixels for $15$ frames, 
and a dense sampling step-size of $5$ pixels, using the code made available by the authors \cite{wang-2011}. 
The features were stored as groups of $4$-byte floating point numbers, 
each associated with a video $v_i$ in dataset $d$, where $i = \{1,...,N\}$. 
With our settings the size of each feature $S_{Gb}$ was $((30_{traj}+96_{HoG}+108_{HoF}+96_{MBHx}+96_{MBHy})\times4)/1024^3 \ {Gb}$, where the summed integers denote the dimensionality of each feature component. 
The approximate number of Dense Trajectory features extracted per dataset and per video is presented in Table \ref{tab:feature_stats}, 
together with further dataset specific statistics. 

\textbf{Visual vocabulary. }
The $k$-means algorithm was used to generate the visual vocabulary for BoF and VLAD vectors. 
The greedy nature of $k$-means means that it may only converge to a local optimum. 
Therefore the algorithm was repeated 8-times to retain the clustering with the lowest error. 
The number of feature samples used to learn the vocabulary was set to a maximum of $10,000$ per cluster. 
We found that setting the pool of feature size to depend on the number of clusters allows faster vocabulary computation for small $K$ without loss of performance. 
The maximum number of features was set to a maximum of $1 \times 10^6$ in order to limit memory usage and computational time for larger $K$.

\textbf{Hardware constraints. } 
With Dense Trajectory features \cite{wang-2011}, 
$1 \times 10^6$ features translate to $\sim$ 1.6Gb of memory when using 32-bit floating point numbers, 
and may thus easily be worked with on a modern PC or laptop. 
All our experiments were carried out on an 8-core, 2GHz, 32GB workstation.


\textbf{Bag-of-features. }
In the BoF approach, 
histograms are computed by quantising features to the nearest cluster (minimum euclidean distance). 
Each BoF histogram was $L1$ normalised separately for each feature component and then jointly \cite{sapienza-2013}. 
The same setup was used for BoF per-category, 
except that separate vocabularies were generated by clustering the features from each class and then joining them into one universal vocabulary. 
For both BoF approaches the exponentiated $\chi^2$ kernel SVM was used: 
\begin{equation}
\mathcal{K}(H_i,H_j) = \exp \left( -\frac{1}{2A} \sum_{n=1}^K \frac{(h_{in} - h_{jn})^2}{h_{in} + h_{jn}} \right), 
\end{equation}
where $H_i$ and $H_j$ are the histograms associated by videos $i$ and $j$, $K$ is the number of cluster centres, 
and $A$ is the mean value of all distances between training histograms \cite{zhang-2007,laptev-2008,wang-2009}.   

\textbf{VLAD and Fisher vectors. }
Initially, due to the high dimensionality of both VLAD and Fisher vectors, 
the Dense Trajectory feature components were independently reduced to 24 dimensions using PCA. 
We used the randomised PCA algorithm by Rokhlin \etal\ \cite{rokhlin-2009} to solve the $USV'$ decomposition on the same $1 \times 10^6$ subsampled features used for dictionary learning. 

Instead of creating a visual vocabulary by clustering the feature space by $k-$means, as done in the BoF and VLAD approach \cite{jegou-2011},
for Fisher vectors it is assumed that the features are distributed according to a parametric Gaussian mixture model (GMM). 
Whereas in BoF, the feature quantisation step is a lossy process \cite{boiman-2008},
a Fisher vector is formed through the soft assignment of each feature point to each Gaussian in the visual vocabulary. 
Therefore, instead of being limited by a hard assignment, 
it encodes additional information about the distribution of each feature \cite{perronnin-eccv-2010}.
Both VLAD and Fisher vectors were computed using the implementations available in the VLFeat library \cite{vedaldi08vlfeat}. 
We followed Perronnin \etal\ and apply power normalisation followed by L2 normalisation to each Fisher vector component separately \cite{perronnin-eccv-2010},
before normalising them jointly.
Contrarily to expectations, it has been shown that Fisher vectors achieve top results with \emph{linear} SVM classifiers \cite{fan-2008,perronnin-eccv-2010}. 

\textbf{Support vector machines. }
For both the $\chi^2$ kernel and linear SVMs, multi-class classification was performed using a 1-vs-all approach \cite{chang-2011}, 
and we keep the regularisation cost C=100 constant throughout \cite{wang-2009,le-2011}.

\subsection{Variable components of the experiments}
\label{subsec:variable_params}
\textbf{Test 1 - comparison of uniform random sampling strategies}. 
We compared two methods of sampling descriptors, 
which formed the data on which the $k$-means algorithm operated in order to learn a visual vocabulary. 
Method \textbf{1a} sampled a balanced random set of features from the dataset, 
accounting for an unbalanced number of videos per action category, 
and an unbalanced number of features per video (see Table \ref{tab:feature_stats}). 
A uniform random selection of features \emph{without} balancing was performed for method \textbf{1b}. 
The exact sampling strategy is detailed in Algorithm \ref{alg:sampling}, whilst the comparative performance will be discussed in section \ref{sec:results}.


\begin{algorithm}[tb]
\caption{The sequence of steps used for generating a visual vocabulary for action classification datasets. 
Note that our proposed method \textbf{(1a)} is designed to be robust to an imbalance in the number of videos per category, 
and to an imbalance in the number of features per video.}
\label{alg:sampling}
\begin{itemize}
\item[1.] Extract Dense Trajectory features \cite{wang-2011} from videos and store them on disk;
\item[2.] For each dataset $d$, calculate the mean number of features $\mu_d = \frac{1}{N} \sum_{i=1}^N n_i$, where $n_i$ is the total number of features extracted from clip $v_i$, and $i = \{1,...,N\}$;
\item[3.] Set the maximum number of videos to sample from $V_{max}$ based on the mean number of features per video: $V_{max} = \lfloor \frac{M_{Gb}}{\mu_d \times S_{Gb}} \rfloor$, 
  where $M_{Gb}$ is the maximum memory available, and $S_{Gb}$ is the memory needed to store a single feature; 
\item[4.] Subsample $\lfloor \frac{V_{max}}{C_d} \rfloor$ videos uniformly at random from each class (\textbf{1a}, class balancing), or $V_{max}$ videos uniformly at random from the entire set of videos (\textbf{1b}), 
  where $C_d$ denotes the number of action categories in dataset $d$;
\item[5.] Load features into memory by uniformly subsampling $\mathrm{min}(n_l, \mu_d)$ features per video (\textbf{1a}), or by loading all the features $n$ in each video (\textbf{1b}), 
where $l \subseteq i$ denotes the set of subsampled video indices;
\item[6.] Subsample $\mathrm{min}(1\times10^6,K\times10^4)$ features (\textbf{1a}) at random, with a balanced number from each class, or (\textbf{1b}) at random from the entire set;
\item[7.] Perform $k$-means with $K$ cluster centres separately per feature component (\textbf{2a}) or jointly by concatenating the feature components (\textbf{2b});
\item[8.] Represent videos globally via BoF (\textbf{3a}), BoF per-category (\textbf{3b}), VLAD vectors (\textbf{3c}), or Fisher vectors (\textbf{3d}).
\end{itemize}
\end{algorithm}

\textbf{Test 2 - comparison of joint and component-wise visual vocabulary generation}. 
Here we learnt a visual vocabulary separately for each feature component (\textbf{2a}) such as the Trajectories, HoG, HoF, MBHx and MBHy components, 
and compared this to grouping these features together and learning a joint vocabulary (\textbf{2b}). 

\textbf{Test 3 - comparison of global representations}. 
For each experimental setting, 
we assessed the performance of the following four global representations: 
standard BoF (\textbf{3a}), BoF per-category (\textbf{3b}), VLAD (\textbf{3c}) and Fisher vectors (\textbf{3d}). 
Each variation was run for 7 values of $K$ cluster centres, 
ranging from $2^2$ to $2^8$ in integer powers of 2.




\section{Results and discussion}
\label{sec:results}
\begin{figure*}[h!]
\centering
\includegraphics[width=1\textwidth]{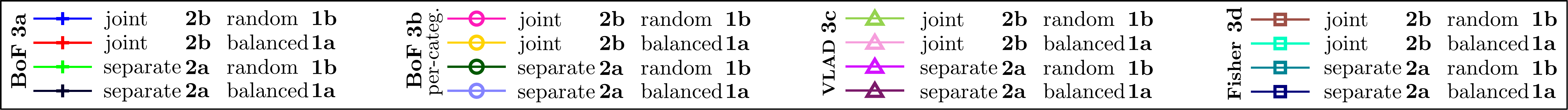}
\vskip 0.2cm
\includegraphics[width=1\textwidth]{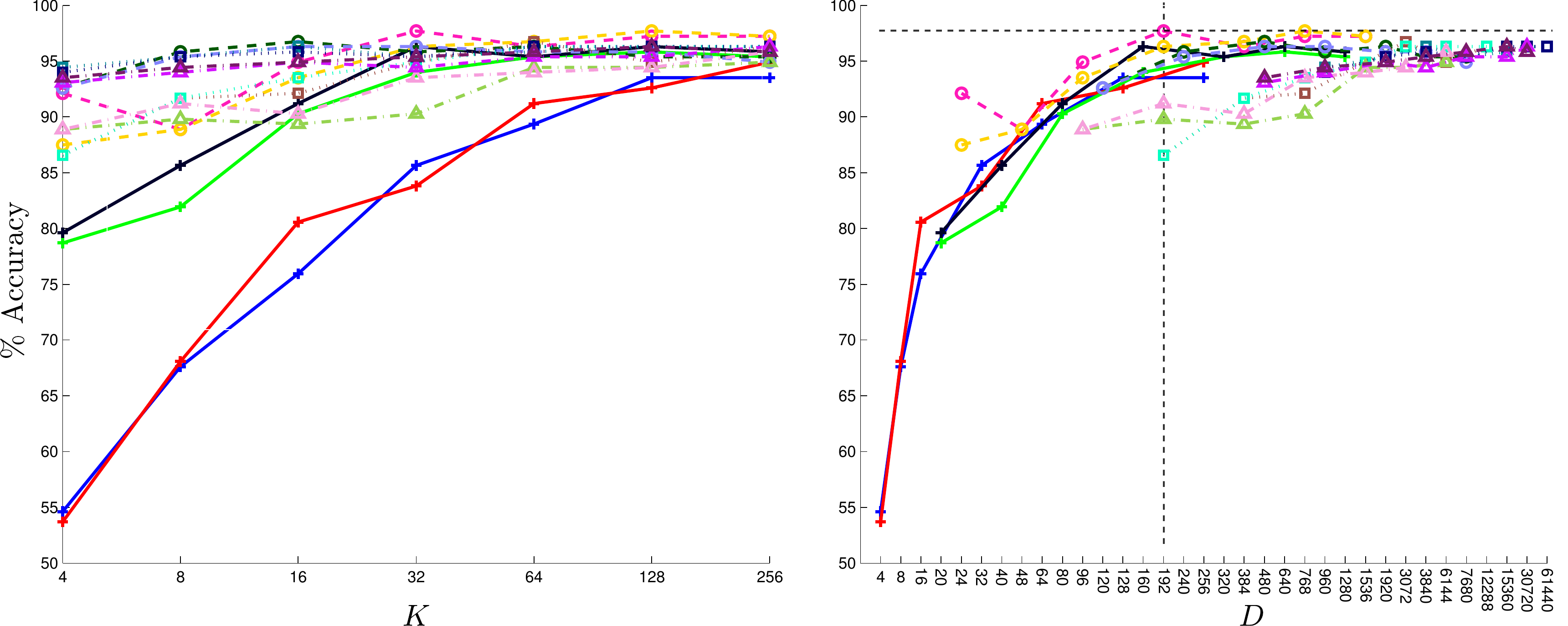}
\vskip -0.3cm
\caption{KTH dataset classification accuracy using a global video representation. 
(left) The Accuracy is plotted against the number of cluster centres $K$, 
and (right) against the representation dimensionality. 
Note that with a small number of cluster centres (left), high dimensional representations may be achieved (right), 
for example when generating vocabularies separately per feature component or per action class. 
In the rightmost plot, at a relatively low representation dimensionality $D=192$, 
there are several competing representations within a $\sim$12\% range in accuracy. 
The best representation method, however, on this dataset is BoF per-category \textbf{(3b)} with a single joint visual vocabulary \textbf{(2b)} and features sampled at random \textbf{(1b)}. 
}
\label{fig:KTH_results}
\end{figure*}

\begin{figure*}[h!]
\centering
\includegraphics[width=1\textwidth]{legend.pdf}
\vskip 0.2cm
\includegraphics[width=1\textwidth]{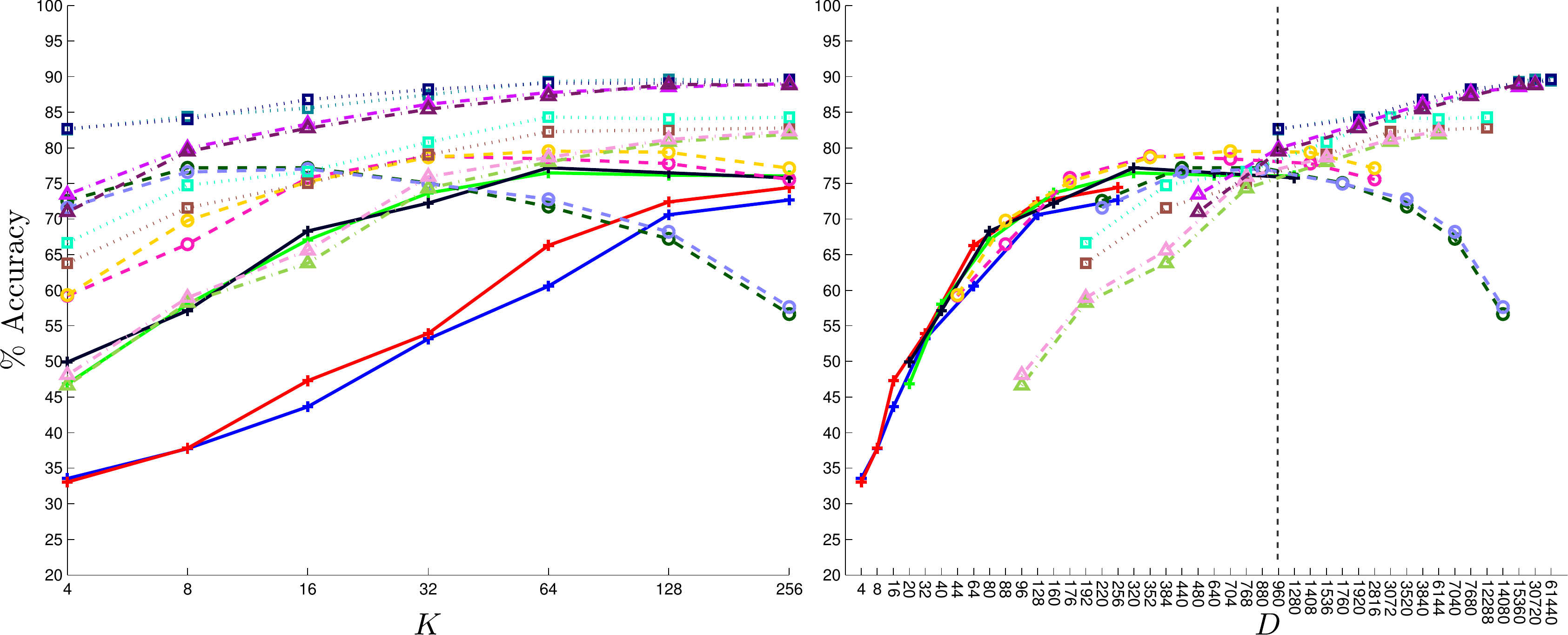}
\vskip -0.3cm
\caption{YouTube dataset classification results. (left) Accuracy vs. $K$ cluster centres, 
and (right) accuracy vs. representation dimensionality. 
Notice (right) the inverted `U' paths traced by the $\chi^2$ kernel SVM (BoF and BoF per category) as opposed to those generated by the
linear SVM (VLAD and Fisher vectors). The dotted line indicates the group of data points at $D=960$ and highlights the top performing method, 
which is Fisher vectors with only 4 clusters per feature component. 
The error bars over 25 folds have been suppressed for clarity. }
\label{fig:YOUTUBE_results}
\end{figure*}

\begin{figure*}[h!]
\centering
\includegraphics[width=1\textwidth]{legend.pdf}
\vskip 0.2cm
\includegraphics[width=1\textwidth]{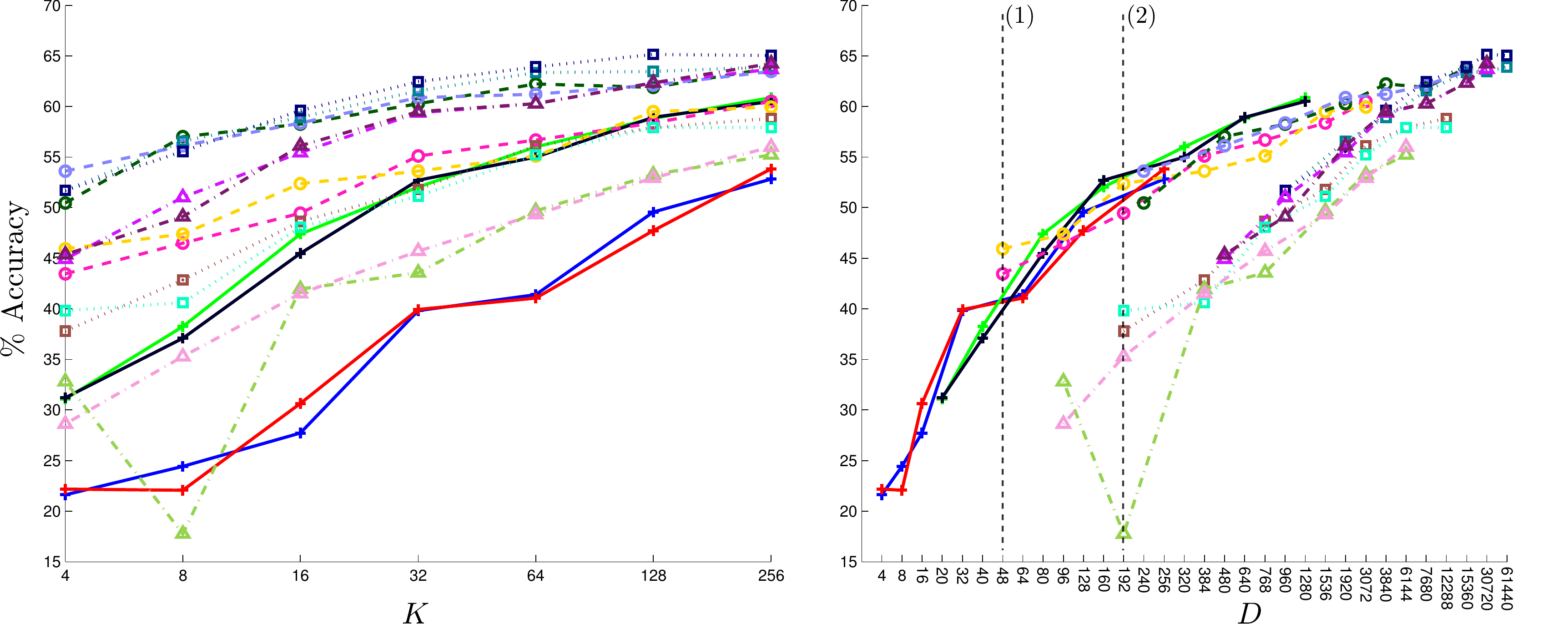}
\vskip -0.3cm
\caption{The classification accuracy versus $K$ cluster centres (left) and representation dimensionality (right) on the Hollywood2 dataset. 
For the BoF method, notice the significant jump in performance (left) between joint (red/blue crosses) and separate (green/black crosses) feature component vocabulary learning. 
It is also noteworthy (right) that at low dimensionalities, 
the Chi-square kernel SVM far outperforms the linear SVM, 
but for higher dimensional vectors, 
linear SVM surpasses the former. 
Dotted lines (1) and (2) highlight situations in which (1) BoF-per category achieves competitive performance for small $K$, 
and (2) random balanced sampling helps preventing situations in which a bias in the sampled feature pool may give poor results. }
\label{fig:HOHA2_results}
\end{figure*}

\begin{figure*}[h!]
\centering
\includegraphics[width=1\textwidth]{legend.pdf}
\vskip 0.2cm
\includegraphics[width=1\textwidth]{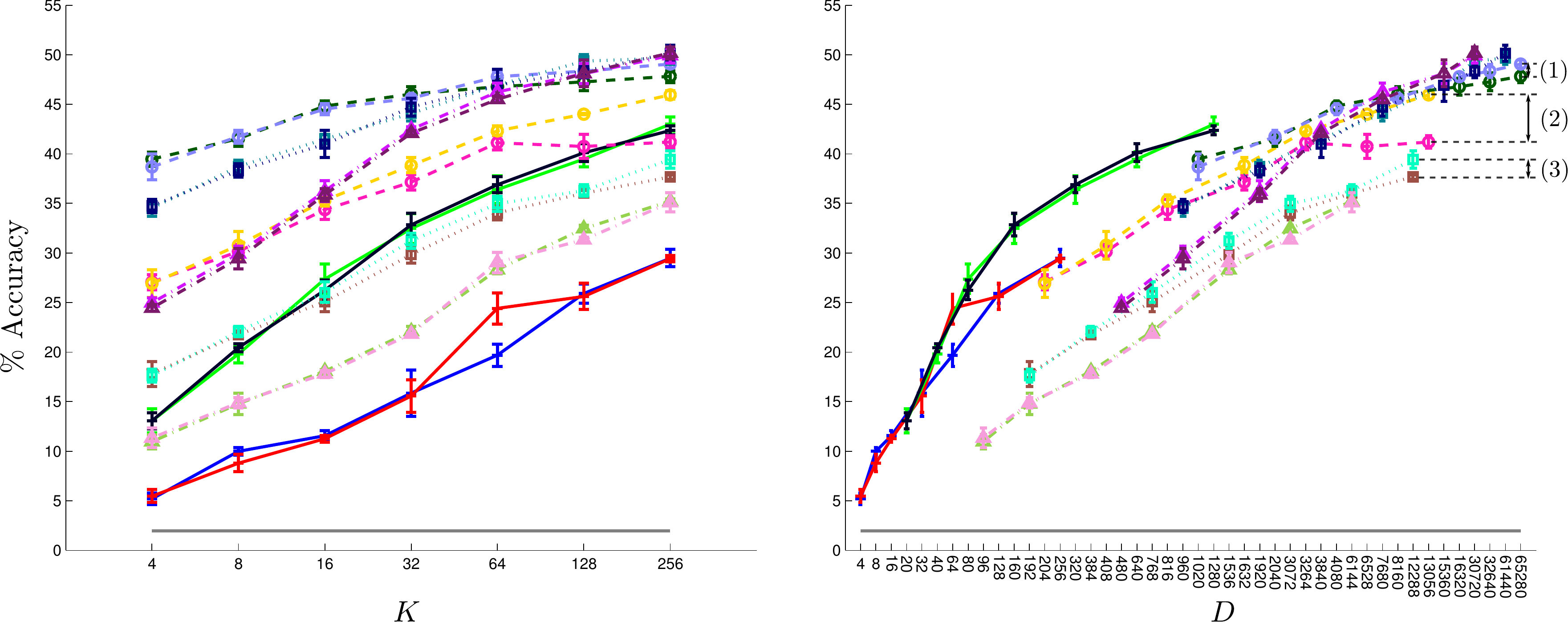}
\vskip -0.3cm
\caption{The classification accuracy versus $K$ cluster centres (left) and representation dimensionality (right) for the HMDB dataset. 
The differences between the random and balanced features becomes more significant at higher values of $K$. 
For example, this is clearly seen between (1) the purple vs. green curves of separate BoF per-category, (2) the yellow vs. pink of joint BoF per-category, and  
(3) the cyan vs. brown curves of Fisher vectors. The error bars in the figure denote one standard deviation over the three train-test splits defined by the HMDB authors \cite{kuehne-2011}. 
The grey horizontal bar just under 2\% marks the chance level. 
}
\label{fig:HMDB_results}
\end{figure*}

Our experiments consisted of 2 binary tests (\textbf{1a-b, 2a-b}) for 7 values of $K$ cluster centres, 4 representation types (\textbf{3a-d}) and 4 datasets, 
for a total of $2^2 \times 7 \times 4 \times 4 = 112 \times 4 = 448$ experimental runs, 
the results of which are shown in Figs.~\ref{fig:KTH_results}-\ref{fig:HMDB_results}. 
The best quantitative results were further listed in Table~\ref{tab:global_results} in order to compare them to the current state-of-the-art.
For each dataset, we presented:
\begin{itemize}
\item[i)] the classification accuracy plotted against the number of clusters $K$, 
\item[ii)] the accuracy plotted against the final representation dimensionality $D$. 
\end{itemize}
Note that the horizontal axis marks for Figs.~\ref{fig:KTH_results}-\ref{fig:HMDB_results} are evenly spaced, thus encoding only the ordering of $K$ and $D$. 
The computational time needed to generate the results was approximately 130 CPU hours for KTH, 1290 CPU hours for YouTube, 1164.8 CPU hours for Hollywood2, and 974 CPU hours on the HMDB dataset.


The KTH dataset results were plotted in Fig.~\ref{fig:KTH_results}.  
The accuracy quickly shot up with increasing dimensionality and saturates. 
By looking at the curves of Fig.~\ref{fig:KTH_results}, 
one may identify the best performing method under a value of $K$ and representation size $D$. For example the dotted line in Fig.~\ref{fig:KTH_results} highlights the methods at $D=192$, 
where several test configurations span an accuracy gap of approximately 12\%. 
In this case, the BoF per-category (\textbf{3b}) at $D=192$ ($32$ clusters $\times 6$ classes) surpassed the current state-of-the-art with 97.69\% accuracy, 98.08\% mAP, and 97.68\% mF1 score. 
Note that the histogram was formed by sampling the features uniformly at random (\textbf{1b}) and clustering the features jointly (\textbf{1b}).

The results from the YouTube dataset did not follow the same pattern as for KTH, as seen in Fig.~\ref{fig:YOUTUBE_results}. 
The inverted `U' pattern generated by the BoF and BoF per-category results indicated that higher dimensional representations, 
when combined with the $\chi^2$ kernel SVM lead to overfitting. 
This was no so for Fisher vectors and linear SVMs, 
which outperformed all the BoF results with only 4 cluster centres per feature component (see dotted line in Fig.~\ref{fig:YOUTUBE_results}-right). 

Here one can clearly see the benefit of using separate vocabularies per feature type (see Fisher joint (\textbf{2b}) vs. Fisher separate (\textbf{2a})), 
and the advantage of random balanced sampling as the number of classes and videos has increased compared to KTH. 
Comparing the results obtained as a function of $K$, 
on the YouTube dataset, Fisher vectors (\textbf{3d}) outperformed the other representation types 96.43\% of the time, 
generating a visual vocabulary separately per feature was best for 85 .71\% of the trials, 
whilst using a random and balanced feature selection was best 66\% of the time. 
%

The Hollywood2 results of Fig.~\ref{fig:HOHA2_results} (right) highlight the initially increased performance obtained using BoF at low values of $K$, 
which were eventually overtaken by the high dimensional Fisher and VLAD vectors with a linear SVM at the top of the curve. 
In fact, at dotted line (1), BoF per-category with joint features and balanced sampling (3b-2b-1a at $D=48$) achieves $5\%$ higher accuracy than the equivalent Fisher vector at $D=192$.
Even though at the top of the curve Fisher vectors achieve the highest accuracy, 
note that BoF per-category is very close behind with a separate (\textbf{2a}) and balanced (\textbf{2a}) vocabulary (purple circle).

The Hollywood2 dataset has the largest variance in the number of features per video clip (see Table \ref{tab:feature_stats}), 
making it more susceptible to randomly selecting an imbalanced feature set for $k$-means. 
For example, the outlier for VLAD at $D=192$ in Fig.~\ref{fig:HOHA2_results} (right, dotted line 2) may be caused by the biased feature sampling of method \textbf{(1b)}, 
which has the undesirable effect of excluding discriminative features from the learned vocabulary, 
and over-representing features which appear more frequently in the dataset. 
Again on this dataset, 
learning vocabularies separately per feature component greatly outperformed learning vocabularies jointly, 
for every representation dimensionality value $D$. 
Here, the best performing approach made use of Fisher vectors (\textbf{3d}), with random-balanced sampling (\textbf{1a}) and features clustered per component (\textbf{2a}), 
which outperformed the current state-of-the-art by 4.32\% (see Table~\ref{tab:global_results}). 
\newcommand{\std}[1]{{\small $\pm${#1}}}
\begin{table}[tb]

\caption{The current state-of-the-art (S-O-T-A) results compared to our best results obtained with the global bag-of-features pipeline (Global best). 
The best results are marked in bold.  
The Variables row indicates which of the varying parameters in Algorithm \ref{alg:sampling} gave the best result. The variations `a' are marked in bold to distinguish them from variations `b'. 
$K-D$ indicates the number of clusters $K$ and the representation dimensionality $D$. 
For datasets with more than one train-test split (YouTube, HMDB, USF51, c.f. Section \ref{sec:datasets}), 
we report one standard deviation around the mean. }

{\setlength{\tabcolsep}{1em}
\begin{center}
\begin{tabular}{| l ||c | c | c | c |}
\hline
\textbf{KTH }   & \textbf{S-O-T-A}      & \textbf{Global best}          & \textbf{Variables}    & \textbf{$K$-$D$}  \\ \hline
Acc         & 96.76 \cite{sapienza-2013}    & \textbf{97.69}            & 3b-2b-1b      & 32 - 192      \\ \hline
mAP         & 97.88 \cite{sapienza-2013}    & \textbf{98.08}            & 3b-2a-1b      & 64 - 384      \\ \hline
mF1     & 96.73 \cite{sapienza-2013}    & \textbf{97.68}            & 3b-2b-1b          & 32 - 192      \\ \hline

\toprule
\hline

\textbf{YouTube }   & \textbf{S-O-T-A}      & \textbf{Global best}      & \textbf{Variables}    & \textbf{$K$-$D$}  \\ \hline
Acc             & 85.4 \cite{wang-2013}     & \textbf{ 89.62 \std{5.41 }}   & 3d-\textbf{2a}-1b     & 128 - 30,720 \\ \hline
mAP             & 89 \cite{oneata-2013}     & \textbf{ 94.25 \std{3.50 }}   & 3d-\textbf{2a}-\textbf{1a} & 128 - 30,720 \\ \hline
mF1         & 82.43 \cite{sapienza-2013}        & \textbf{ 87.45 \std{6.92 }}   & 3d-\textbf{2a}-1b & 128 - 30,720 \\ \hline

\toprule
\hline

\textbf{Hollywood2 }    & \textbf{S-O-T-A}      & \textbf{Global best}      & \textbf{Variables}    & \textbf{$K$-$D$}  \\ \hline
Acc             & 60.85 \cite{sapienza-2013}    & \textbf{65.16}    & 3d-\textbf{2a}-\textbf{1a}  & 128 - 30,720 \\ \hline
mAP             & \textbf{63.3} \cite{oneata-2013}      & 59.66                 & 3d-\textbf{2a}-\textbf{1a}  & 256 - 61,440 \\ \hline
mF1             & 52.03 \cite{sapienza-2013}    & \textbf{54.55}        & 3d-\textbf{2a}-\textbf{1a} & 256 - 61,440 \\ \hline

\toprule
\hline

\textbf{HMDB }      & \textbf{S-O-T-A}      & \textbf{Global best}          & \textbf{Variables}        & \textbf{$K$-$D$}  \\ \hline
Acc             & 48.3 \cite{wang-2013}     & \textbf{50.17 \std{0.614}}        & 3c-\textbf{2a}-\textbf{1a} & 256 - 61,440 \\ \hline
mAP             & \textbf{54.8} \cite{oneata-2013}  & 50.07 \std{0.33}      & 3d-\textbf{2a}-\textbf{1a} & 256 - 61,440\\ \hline
mF1             & \textbf{52.03} \cite{sapienza-2013}   & 48.88 \std{0.94}      & 3d-\textbf{2a}-\textbf{1a} & 256 - 61,440 \\ \hline

\toprule
\hline

\textbf{USF101 }    & \textbf{ S-O-T-A }        & \textbf{ Global best }        & \textbf{ Variables }          & \textbf{ $K$-$D$ }    \\ \hline
Acc             & 43.9 \cite{soomro-2012}, \textbf{85.9}\cite{wang-2013-iccvworkshop}   & 81.24 \std{1.11}          & 3d-\textbf{2a}-\textbf{1a}    & 256 - 61,440\\ \hline
mAP             & --            & \textbf{82.35 \std{0.97}}             & 3d-\textbf{2a}-\textbf{1a}    & 256 - 61,440 \\ \hline
mF1             & --            & \textbf{80.57 \std{1.11}}             & 3d-\textbf{2a}-\textbf{1a}    & 256 - 61,440\\ \hline
\toprule
\end{tabular}
\end{center}}
\label{tab:global_results}
\end{table}

Large performance differences were also observed in Fig.~\ref{fig:HMDB_results} for the HMDB dataset. 
With an increased number of action classes, 
one can more easily see the performance gained by `balanced' sampling rather than uniform random sampling. 
For example, see the differences between the yellow and pink curves of BoF joint per-category (Fig.~\ref{fig:HMDB_results} right, dotted lines 2), 
or the cyan and brown curves of Fisher vector joint random vs. random-balanced (Fig.~\ref{fig:HMDB_results} right, dotted lines 3). 
This was only observed at higher dimensionalities though, and overall `balanced' sampling outperformed random sampling just 53\% of the time. 
Note however that the best accuracy, mean average precision and mean F1 scores on the Hollywood2 and HMDB datasets were obtained using random-balanced sampling, 
as shown in Table~\ref{tab:global_results}. 
In Fig.~\ref{fig:HMDB_results}, one can also observe the benefit of BoF per-category, 
which outperforms BoF by learning separate vocabularies per action class. 

We estimated that to run an additional 112 experiments on UCF101 dataset would take approximately 1 CPU year, 
and thus report only a single run with the best performing variable components observed in the experiments so far, 
namely Fisher vectors, balanced sampling and separate vocabulary learning. 
UCF101 is the largest current action recognition dataset (see Table~\ref{tab:feature_stats}), however it may not be the most challenging.  
We achieved, 81.24\%, 82.35\%, 80.57\% accuracy, mAP and F1 scores respectively averaged over the three train test splits provided by the authors \cite{soomro-2012}. 
Our reported accuracy is 37.34\% higher than the original results reported in \cite{soomro-2012}, and 4.6\% short of the result reported in \cite{wang-2013}, 
which additionally used spatial pyramids and a variation of the Dense Trajectory features which estimates camera motion information \cite{wang-2013-iccv}. 

The computational time needed to generate this result on the UCF101 results was 163.52 CPU hours. 
Interestingly, a large part of the computational time is spent loading features from disk.
Thus even though the chance level of UCF101 is just under 1\%, our results indicate that the HMDB dataset remains the most challenging action classification dataset currently available.  
The overall best results are listed in Table~\ref{tab:global_results}; 
by using our approach (Algorithm \ref{alg:sampling}) with an increased number of model parameters (using a larger K, or by using spatial pyramids\cite{lazebnik-2006}) may yield further improvements to the current state-of-the-art results.

\section{Conclusion}
In this work we proposed to evaluate aspects of the bag-of-features pipeline previously unexplored, 
on the largest and most challenging action recognition benchmarks, and achieved state-of-the-art results across the board. 
In particular, we focused on the feature subsampling and partitioning step after features have been extracted from videos, but prior to encoding. 
Since uniform random sampling of features to create a vocabulary may be biased towards longer videos and action classes with more data, 
we compared uniform random sampling to sampling a random and balanced subset from each video and action class. 
The best results obtained showed that the proposed sampling strategies did yield minor performance improvements, 
and helped preventing poor results that may follow from sampling an unbalanced set of features (see Fig. \ref{fig:HOHA2_results} dotted line 2); 
other balancing strategies may provide further improvements. 
We also showed that learning separate vocabularies per feature component caused very large improvements in performance. 
Furthermore, learning BoF per-category outperformed BoF, 
but was surpassed by Fisher vectors on the larger and more challenging datasets. 
Finally, our results demonstrated that competitive results may be achieved by using $k$-means with a relatively small number of cluster centres $K$.

\bibliographystyle{splncs}

\bibliography{references}

\end{document}